\documentclass[letterpaper, 10 pt, conference]{ieeeconf}

\IEEEoverridecommandlockouts

\usepackage[OT1]{fontenc}
\usepackage{graphics}
\usepackage{epsfig}
\usepackage{mathptmx}
\usepackage{times}
\usepackage{cite}
\usepackage{amsmath,amssymb,amsfonts}
\usepackage{algorithmic}
\usepackage{graphicx}
\usepackage{placeins}
\usepackage{booktabs}
\usepackage{textcomp}
\usepackage{xcolor}
\usepackage{flushend}
\usepackage{multirow}

\title{\LARGE \bf
GR-VIP: Geometric Representation Conditioning for Precise Visuomotor Policy
}

\author{
Chanhwi Lee$^{1,*}$, Jiwon Seong$^{1,*}$, Eunje Choi$^{1}$, Hohyung Lee$^{1}$, Jongsoo Lee$^{1}$,\\
Junyeong Bae$^{1}$, Inveom Kwak$^{2}$,
Woosang Shin$^{3,\dagger}$, and Soohee Han$^{1,\dagger}$
\thanks{$^{*}$Equal contribution; $^{\dagger}$Corresponding authors.}%
\thanks{$^{1}$Chanhwi Lee, Jiwon Seong, Eunje Choi, Hohyung Lee, Jongsoo Lee, Junyeong Bae, and Soohee Han are with Pohang University of Science and Technology (POSTECH) (e-mail: \texttt{soohee.han@postech.ac.kr}).}%
\thanks{$^{2}$Inveom Kwak is with Polaris3D
(e-mail: \texttt{klauski@polaris3d.co}).}%
\thanks{$^{3}$Woosang Shin is with the Korea Institute of Industrial Technology (KITECH) (e-mail: \texttt{welldone@kitech.re.kr}).}%
\thanks{This work was supported by the Technology Development Program (RS-2023-00321740), funded by the Ministry of SMEs and Startups (MSS), Korea.}
\thanks{This work has been submitted to the IEEE for possible publication. Copyright may be transferred without notice, after which this version may no longer be accessible.}
}

\begin{document}

\maketitle
\raggedbottom
\thispagestyle{empty}
\pagestyle{empty}

\begin{abstract}
Achieving precise terminal position and heading alignment remains a significant challenge for learning-based mobile robot control. Recent vision-based learning methods have shown promise for precise target-relative control and generalization to unseen environments with pretrained visual encoders, but general-purpose visual representations are not specifically designed to capture the fine-grained spatial differences between current and goal views required for precise control, and real-world demonstrations combining millimeter-scale terminal accuracy with zero-shot transfer across environments remain limited. To address these challenges, we propose Geometric Representation Conditioning for Precise Visuomotor Policy (GR-VIP), a goal-conditioned visuomotor policy that leverages a visual encoder pretrained for relative camera pose estimation. Rather than using the final relative-pose estimate, GR-VIP conditions action prediction on the dense relational features produced before pose regression, preserving fine-grained cross-view information for learning corrective actions from demonstrations. We evaluate GR-VIP on a real-world docking task and demonstrate high docking success rates and millimeter-scale terminal accuracy in both the training and unseen environments, without additional training for transfer. Furthermore, supplementary experiments illustrate the policy's ability to align with non-docking targets and complete docking under a tested low-light condition.
\end{abstract}


\section{Introduction}
Learning-based methods enable mobile robot behaviors to be learned directly from data, rather than relying on manually designed control strategies~\cite{argall2009survey,ravichandar2020recent}.
These approaches have been applied to a wide range of mobile robot tasks, including navigation~\cite{tai2017virtual,cesar2021improving}, trajectory tracking~\cite{zhang2019trajectory,da2023accurate}, and visual servoing~\cite{jin2021gaussian,jin2021policy}. However, achieving precise terminal alignment remains challenging in applications such as docking, inspection, and manipulation-ready base positioning, where successful operation requires accurate positioning and orientation relative to a target~\cite{burgueno2021autonomous,meng2025aim,yang2025mobi}.

Recent vision-based learning methods have explored precise target-relative positioning for mobile robots~\cite{meng2025aim,lee2026learning}. These approaches have demonstrated promising performance in precise positioning and generalization to unseen objects or conditions. However, precise control requires the policy to distinguish fine-grained spatial differences between the current and desired configurations and translate them into appropriate corrective actions. General-purpose visual representations are not explicitly pretrained to encode goal-relative geometric relationships, and existing approaches often introduce additional geometric or target-specific processing to support precise control.
Despite these advances, real-world demonstrations that combine millimeter-scale terminal accuracy with zero-shot transfer across environments remain limited.

To address these challenges, we propose Geometric Representation Conditioning for Precise Visuomotor Policy (GR-VIP), a goal-conditioned visuomotor policy that uses a visual encoder pretrained for relative camera pose estimation. We hypothesize that geometry-aware cross-view representations provide a suitable prior for precise goal-relative control, rather than relying on the final relative-pose estimate. GR-VIP therefore retains dense relational features before pose regression and uses them for policy conditioning. This allows the policy to exploit fine-grained geometric information while learning appropriate corrective actions from demonstrations.

\begin{figure*}[!t]
    \centering
    \includegraphics[width=\textwidth]{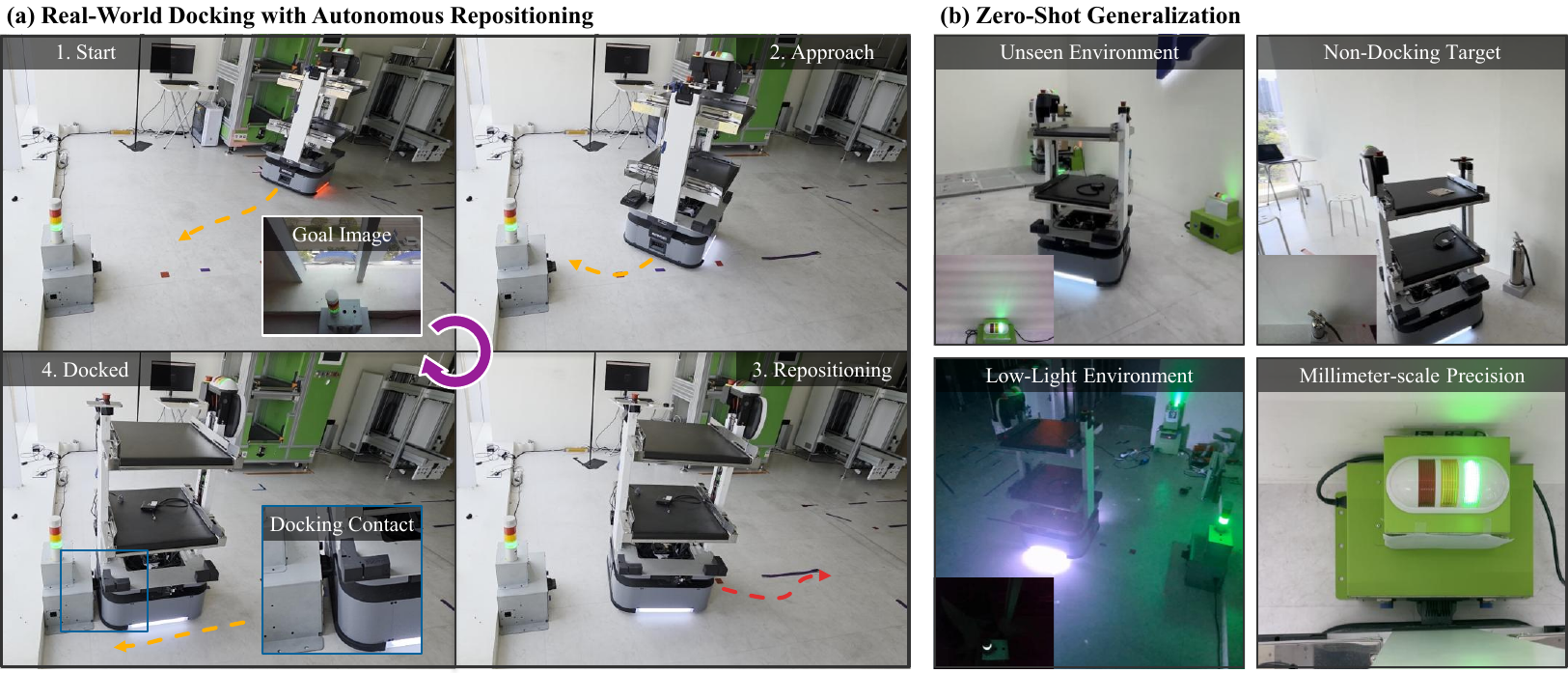}
    \caption{Overview of GR-VIP.
    (a) GR-VIP performs precise real-world docking with autonomous repositioning to correct misalignment.
    (b) 
    A policy trained in a single environment transfers to unseen environments without additional training. The same policy also demonstrates alignment to non-docking targets and successful docking under a low-light condition.}
    \label{fig:teaser}
\end{figure*}

We evaluate GR-VIP on a real-world mobile robot docking task requiring precise terminal position and heading alignment. We compare our method with baselines using alternative visual representations in the training environment and under unseen background and target conditions. 
Our experiments show that GR-VIP achieves millimeter-scale terminal accuracy and a high docking success rate in the training environment, while maintaining high success rates and millimeter-scale accuracy in unseen environments without additional training.
Furthermore, additional experiments illustrate alignment to configurations specified by non-docking goal images and successful docking under a low-light condition. Fig.~\ref{fig:teaser} provides an overview of GR-VIP's capabilities.

\section{Related Work}
\label{sec:related_work}

\subsection{Precise Docking Systems}
Precise mobile robot docking requires accurate terminal position and heading alignment for successful physical engagement~\cite{yilmaz2024multi,wang2020autonomous}. A common approach is to estimate the relative pose between the robot and the docking target and use this estimate to control the final approach and alignment~\cite{wang2020autonomous}. Marker-based methods use fiducial markers installed on the docking target as visual references to estimate relative position and orientation for precise docking~\cite{bolanakis2021qr,chinchilla2024real}. Laser-based methods have also been proposed to detect charging stations and guide the docking process~\cite{liu2022laser}. 
Registration-based approaches instead exploit the geometric structure of the target directly. Gutierrez et al.~\cite{gutierrez2025real} extract line features from sparse point clouds and use the geometry of the docking station for global registration, estimating the poses of both the robot and the station with millimeter-level positioning accuracy. Although these approaches can achieve precise docking, repositioning maneuvers to correct misalignment are less commonly addressed. In contrast, GR-VIP is trained on demonstrations that include such maneuvers, enabling the policy to learn repositioning behavior directly from data.

\subsection{Goal-Conditioned Navigation for Precise Local Alignment}

Image-goal navigation provides a flexible formulation in which a destination is specified by a goal image rather than accurate target information~\cite{chaplot2020neural,wu2022image}. ViNT~\cite{shah2023vint} and NoMaD~\cite{sridhar2024nomad} learn transferable goal-conditioned navigation policies from data collected across diverse robots and environments. However, these methods primarily focus on long-horizon goal reaching and do not directly address the precise terminal position and heading alignment required for physical docking. More recent work has explored spatial relationships between current and goal observations for precise local alignment.
AnyImageNav~\cite{deng2026anyimagenav} adopts a learning-free approach that treats the goal image as a geometric query, matches it against accumulated observations, and recovers the 6-DoF pose of the goal viewpoint for navigation. Lee et al.~\cite{lee2026learning} ground the target object and use object-centric spatial correlations between the current and goal views for manipulation-ready positioning. Aim My Robot~\cite{meng2025aim} explicitly specifies the desired spatial relation as a target-relative base pose and predicts a trajectory toward that pose using RGB-D and LiDAR observations. While these approaches advance precise local alignment, they often incorporate additional sensing or explicit target-related processing beyond the goal image. Moreover, the combination of millimeter-scale terminal accuracy and zero-shot transfer across real-world environments remains underexplored. In contrast, GR-VIP performs precise local alignment using monocular RGB observations and proprioceptive information, while achieving millimeter-scale accuracy and transfer to unseen environments without additional training.

\begin{figure*}[!t]
\centering
\includegraphics[width=0.90\textwidth]{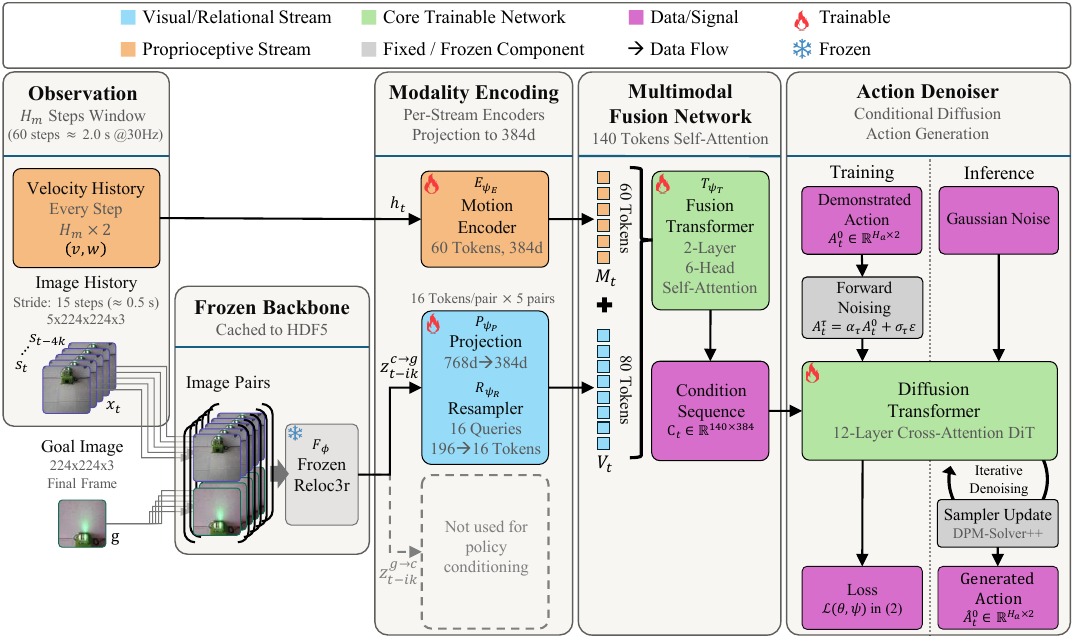}
\caption{GR-VIP architecture. Dense relational tokens from the frozen Reloc3r capture geometric relationships between each observation and the goal image. The tokens are resampled and fused with wheel-velocity history to condition a diffusion policy that predicts future velocity commands.}
\label{fig:GRVIP_architecture}
\end{figure*}

\subsection{Visual Representations for Visuomotor Control}
In visuomotor control, the representation used to encode visual observations can substantially affect downstream control performance~\cite{chen2020robust,majumdar2023we,burns2023makes}. Accordingly, a wide range of visual representations have been explored for visuomotor policies. 
R3M~\cite{nair2022r3m} transfers representations pretrained on diverse human videos to manipulation policies. MVP~\cite{xiao2022masked} uses masked visual pretraining on large-scale real-world images to provide representations for downstream motor control.
General-purpose visual representations such as DINOv2 have also been adopted in visuomotor policies~\cite{kim2024openvla}.
More recent approaches incorporate spatial or geometric information into visuomotor policies. DP3~\cite{ze20243d} conditions a diffusion policy on 3D representations extracted from sparse point clouds, while VGGT-DP~\cite{ge2025vggt} integrates geometric priors from a pretrained 3D perception model with proprioceptive information. In contrast, GR-VIP uses dense relational features extracted from current–goal image pairs by an encoder pretrained for relative camera pose estimation, providing a pairwise representation tailored to goal-relative control.

\section{Proposed Method}
\label{sec:method}

We propose GR-VIP, a goal-conditioned visuomotor policy for precise terminal alignment. Given a sequence of recent observations and a goal image specifying the desired configuration, the policy extracts goal-relative relational features using a pretrained visual encoder. These visual features are fused with proprioceptive information derived from wheel-velocity history to condition a diffusion-based model that generates a sequence of control actions. As illustrated in Fig.~\ref{fig:GRVIP_architecture}, the overall framework consists of goal-relative feature extraction, multimodal token encoding and fusion, and diffusion-based action generation.

\subsection{Goal-Relative Geometric Representation}
\label{subsec:geometric_representation}

We first describe how GR-VIP extracts goal-relative relational features from the observation history and goal image. At timestep \(t\), GR-VIP receives a visual history $x_t = \{s_{t-ik}\}_{i=4,\cdots,0}$ consisting of multiple RGB images, along with a goal image \(g\) specifying the desired terminal configuration. Here, $s$ is a visual observation, while $k$ represents the temporal stride between consecutive observations. 

To obtain goal-relative relational features, each observation in $x_t$ is paired with the goal image $g$ and processed by a frozen Reloc3r~\cite{reloc3r}, a large-scale pretrained model for generalizable relative camera pose estimation. It first encodes the two images independently into patch-level feature tokens using a shared Vision Transformer (ViT) encoder. Its symmetric decoder then exchanges information between the two feature streams through cross-attention, producing dense tokens that encode the relationship between the two views. 
We denote the mapping from an observation–goal pair to these decoded relational tokens by \(\mathcal{F}_{\phi}\). Formally, the resulting relational tokens are given by
\begin{equation}
\nonumber
\mathcal{F}_{\phi}(s_{t-ik}, g) = \left(z_{t-ik}^{c\rightarrow g}, z_{t-ik}^{g\rightarrow c} \right),
\end{equation} yielding
\begin{equation} 
F_{\phi}(x_t,g) = \left\{ \mathcal{F}_{\phi}\!\left(s_{t-ik},g\right) \right\}_{i=4,\ldots,0}, \label{sequence-of-token}
\end{equation} where the parameters \(\phi\) remain frozen during policy
training, and \(z^{c\rightarrow g}\) and
\(z^{g\rightarrow c}\) denote the observation-to-goal
and goal-to-observation relational feature streams, respectively. Reloc3r further feeds the decoded tokens into a pose regression head to estimate the relative camera pose. In contrast, GR-VIP directly uses the dense decoded tokens for policy conditioning, bypassing the final pose prediction. Since the policy predicts goal-directed actions from the current observation, we retain only the observation-to-goal feature stream \(z^{c\rightarrow g}\).

\subsection{Multimodal Token Encoding and Fusion}
\label{subsec:multimodal_fusion}

The extracted relational features are fused with the wheel-velocity history through a Transformer-based network. The sequence of decoded relational tokens in Eq.~\eqref{sequence-of-token} is first projected
into the policy embedding space and then compressed into a sequence of visual tokens using a query-based resampler with cross-attention~\cite{jaegle2021perceiver} as

\begin{equation}
\nonumber
V_t = \left\{(R_{\psi_R} \circ P_{\psi_P})\left(z_{t-ik}^{c\rightarrow g}\right)\right\}_{i=4,\ldots,0},
\end{equation}
where \(P_{\psi_P}\) is a linear projection and \(R_{\psi_R}\) is a query-based resampler.
Here, temporal and slot embeddings are added to each resampled visual token to encode its corresponding observation time and resampler output index, respectively.

The velocity history
\(h_t=\{(v_{t-i},w_{t-i})\}_{i=H_m-1,\ldots,0}\), comprising linear and angular velocities over \(H_m\) timesteps, is encoded into a sequence of motion tokens as
\begin{equation}
\nonumber
M_t = E_{\psi_E}(h_t),
\end{equation}
where \(E_{\psi_E}\) denotes the motion encoder.
Temporal embeddings are then used to encode the order of the motion tokens within the velocity history. Before fusion, modality embeddings are added to distinguish the visual and motion token streams.

The resulting visual and motion tokens are concatenated and jointly processed by a Transformer fusion network through self-attention as
\begin{equation}
\nonumber
C_t = T_{\psi_T}\left([V_t;M_t]\right),
\end{equation}
where \(T_{\psi_T}\) denotes the fusion Transformer and \(C_t\) is the multimodal condition sequence.
The resulting condition sequence is subsequently used for action generation. Through this multimodal fusion pipeline, the policy can capture dependencies between the robot's recent motion and temporal
changes in the cross-view cues encoded by the visual tokens. For notational convenience, we collectively denote the learnable parameters introduced above by \(\psi=\{\psi_P,\psi_R,\psi_E,\psi_T\}\).

\subsection{Diffusion-Based Action Generation}
\label{subsec:diffusion-based}

Given the multimodal condition sequence \(C_t\), GR-VIP generates a future
sequence of linear and angular velocity commands using a conditional diffusion
model. Such models are well suited to modeling multimodal action
distributions~\cite{chi2025diffusion}, allowing the policy to capture multiple
valid control behaviors present in the demonstrations.
Formally, the normalized action sequence at timestep $t$ is defined by
\begin{equation}
\nonumber
A_t^0 = \{a_{t+i}\}_{i=0,\cdots,H_a-1},
\end{equation}
where \(a_{t+i}=(v_{t+i}, w_{t+i})\), and \(H_a\) denotes the action horizon. During training, the demonstration action sequence is perturbed at diffusion step \(\tau\) as
\begin{equation}
\nonumber
A_t^\tau = \alpha_\tau A_t^0 + \sigma_\tau\epsilon,\qquad\epsilon\sim\mathcal{N}(0,I),
\end{equation}
where \(\alpha_\tau\) and \(\sigma_\tau\) specify the signal and noise levels under a cosine noise schedule~\cite{nichol2021improved}, satisfying \(\alpha_\tau^2+\sigma_\tau^2=1\).

A Transformer-based diffusion denoiser predicts the noise added to \(A_t^0\). We employ a diffusion Transformer (DiT)~\cite{peebles2023scalable}, parameterized by \(\theta\), where \(C_t\) is incorporated through cross-attention. The model is trained with the noise-prediction objective defined by 
\begin{equation}
\mathcal{L}(\theta,\psi) = \mathbb{E}_{(x_t,h_t,A_t^0),\,\tau,\,\epsilon}\left[\left\|\epsilon -
\epsilon_\theta\left(A_t^\tau,\tau,C_t\right)\right\|_2^2\right],
\end{equation}
where \(\epsilon_\theta\) denotes the noise predicted by the diffusion denoiser. 
All trainable components downstream of the pretrained geometric feature extractor are optimized jointly. During training, we maintain exponential-moving-average (EMA) copies of the conditioning network and diffusion denoiser parameters, which are used for inference.

\begin{figure}[!htb]
    \centering
    \includegraphics[width=\columnwidth]{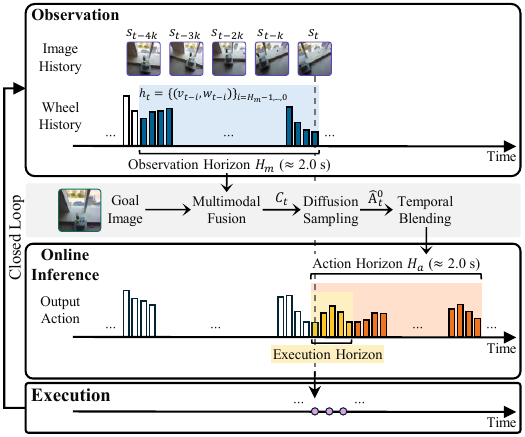}
    \caption{Closed-loop execution of GR-VIP. Multimodal observations condition diffusion-based action generation. Successive action plans are blended to maintain temporal continuity and executed in a receding-horizon manner as new predictions become available. Here, $H_m$ and $H_a$ are both set to 60 steps.
    }
    \label{fig:inference_architecture_process}
\end{figure}

\subsection{Online Inference and Closed-Loop Execution}
\label{subsec:closed_loop}

At inference time, GR-VIP generates a normalized action sequence \(\widehat{A}_t^0\) from Gaussian noise using DPM-Solver++(2M)~\cite{lu2025dpm}, conditioned on \(C_t\), and then denormalizes it to physical velocities. Following a receding-horizon execution scheme~\cite{kwon2005receding,chi2025diffusion}, only a short prefix is executed at each update while new action sequences are generated from updated observations. To promote temporal continuity across successive predictions, the denormalized sequence is blended with the temporally aligned portion of the previously generated sequence before execution.  As illustrated in Fig.~\ref{fig:inference_architecture_process}, the action plan is repeatedly updated as new observations become available.

\begin{figure*}[!t]
\centering
\includegraphics[width=0.96\textwidth]{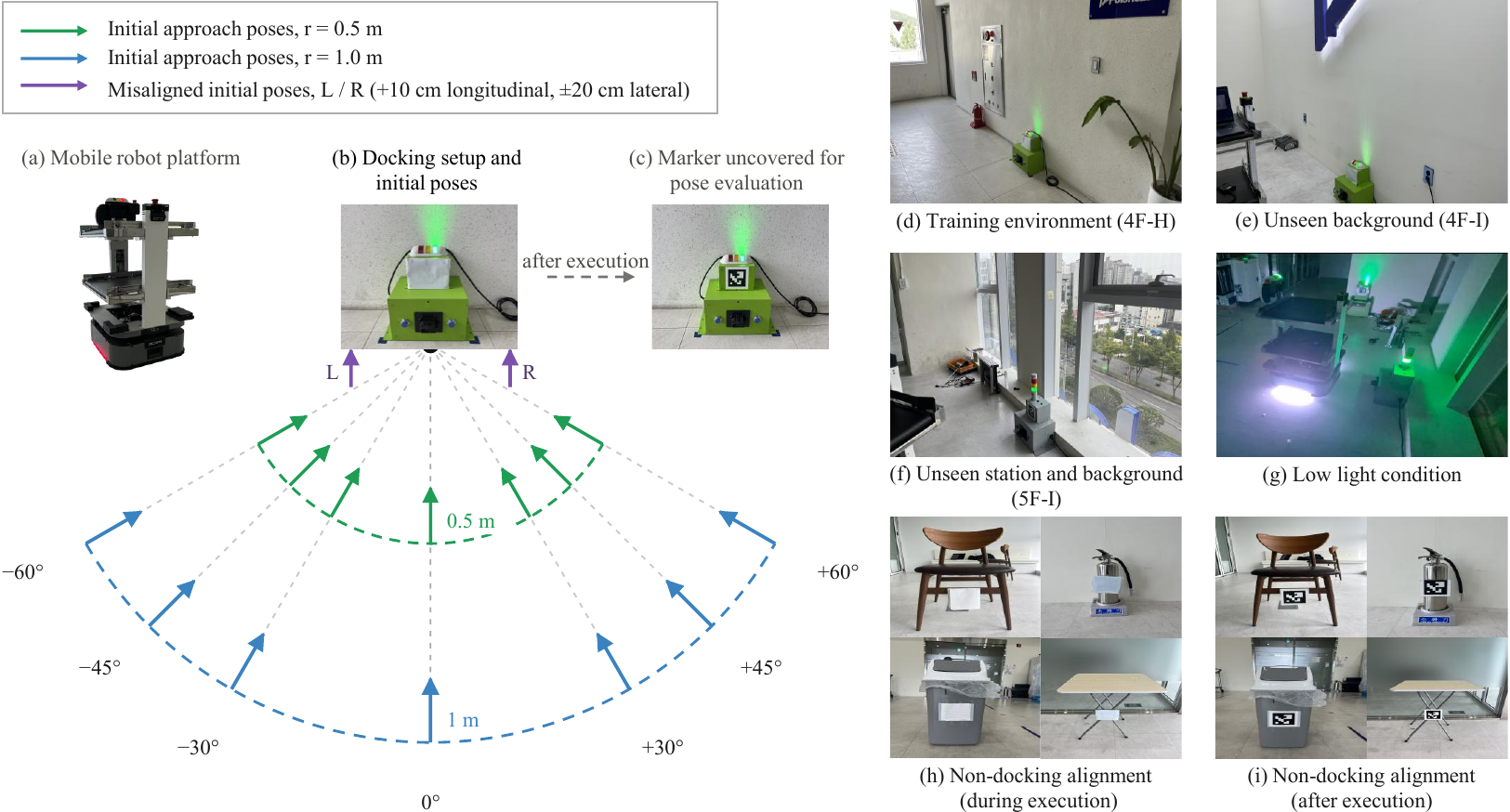}
\caption{Experimental setup. (a) Mobile robot platform. (b) Docking setup and initial poses. Dashed arcs indicate 0.5- and 1.0-m approach distances, and purple arrows denote intentionally misaligned initial poses. (c) Marker uncovered for terminal-pose evaluation after policy execution. (d) Training environment (4F-H). (e) Unseen-background environment (4F-I). (f) Environment with an unseen docking station and background (5F-I). (g) Low-light condition. (h, i) Alignment to a non-docking target during policy execution and subsequent terminal-pose evaluation.}
\label{fig:docking_scenarios}
\end{figure*}

\section{Experimental Evaluation}
\label{sec:experiments}

The experiments evaluate GR-VIP from four perspectives: 1) physical docking success and terminal precision, 2) zero-shot transfer to unseen environments and docking stations, 3) the effect of representation choice on policy conditioning, and 4) supplementary evaluations of non-docking target alignment and docking under a low-light condition.

\begin{table*}[!t]
    \centering
    \caption{Quantitative evaluation of docking performance, zero-shot transfer, representation ablations, and alignment to non-docking targets.}
    \label{tab:docking_results}

    \footnotesize
    \setlength{\tabcolsep}{2pt}
    \renewcommand{\arraystretch}{1.05}

    \begin{tabular*}{\textwidth}{@{\extracolsep{\fill}}lccccccc@{}}
        \toprule
        Model & Train & Eval setting & Success \% (count) & $n/N$ & Lateral (mm) & Longitudinal (mm) & Heading ($^\circ$) \\
        \midrule

        \multicolumn{8}{c}{(a) Training-environment docking performance} \\
        \midrule
        GR-VIP & \multirow{3}{*}{4F-H} & \multirow{3}{*}{4F-H} & \textbf{100.0 (16/16)} & 16/16 & $\mathbf{2.53} \pm 2.24$ & $\mathbf{1.95} \pm 1.41$ & $2.28 \pm 1.64$ \\
        GC-DP (NoMaD adapted) & & & 56.3 (9/16) & 13/16 & $9.44 \pm 6.23$ & $9.86 \pm 10.62$ & $2.09 \pm 1.17$ \\
        DINOv3-based & & & 50.0 (8/16) & 15/16 & $12.79 \pm 10.40$ & $12.66 \pm 12.63$ & $\mathbf{1.81} \pm 2.11$ \\
        \midrule

        \multicolumn{8}{c}{(b) Zero-shot transfer and environment-specific baselines} \\
        \midrule
        GR-VIP & \multirow{3}{*}{4F-H} & \multirow{3}{*}{4F-I} & \textbf{100.0 (16/16)} & 16/16 & $7.20 \pm 4.78$ & $7.15 \pm 4.86$ & $3.75 \pm 1.71$ \\
        GC-DP (NoMaD adapted) & & & 0.0 (0/16) & 2/16 & $130.88 \pm 13.59$ & $35.26 \pm 1.53$ & $38.08 \pm 0.92$ \\
        DINOv3-based & & & 31.3 (5/16) & 11/16 & $15.42 \pm 15.16$ & $19.54 \pm 10.08$ & $10.18 \pm 3.04$ \\
        \midrule
        GC-DP (NoMaD adapted) & \multirow{2}{*}{4F-I} & \multirow{2}{*}{4F-I} & 31.3 (5/16) & 6/16 & $\mathbf{6.47} \pm 6.13$ & $10.42 \pm 4.99$ & $\mathbf{2.19} \pm 1.53$ \\
        DINOv3-based & & & 50.0 (8/16) & 15/16 & $21.29 \pm 29.16$ & $11.77 \pm 7.90$ & $4.24 \pm 2.83$ \\
        \midrule
        GR-VIP & \multirow{3}{*}{4F-H} & \multirow{3}{*}{5F-I} & \textbf{93.8 (15/16)} & 16/16 & $\mathbf{2.21} \pm 1.93$ & $\mathbf{5.62} \pm 2.85$ & $4.69 \pm 2.13$ \\
        GC-DP (NoMaD adapted) & & & 0.0 (0/16) & 0/16 & -- & -- & -- \\
        DINOv3-based & & & 0.0 (0/16) & 0/16 & -- & -- & -- \\
        \midrule
        GC-DP (NoMaD adapted) & \multirow{2}{*}{5F-I} & \multirow{2}{*}{5F-I} & 87.5 (14/16) & 15/16 & $3.62 \pm 2.57$ & $6.83 \pm 5.86$ & $\mathbf{1.20} \pm 0.80$ \\
        DINOv3-based & & & 81.3 (13/16) & 16/16 & $9.38 \pm 12.71$ & $11.38 \pm 11.06$ & $1.54 \pm 1.07$ \\
        \midrule

        \multicolumn{8}{c}{(c) Representation ablation for policy conditioning} \\
        \midrule
        \multirow{3}{*}{Reloc3r pre-decoder features} & \multirow{3}{*}{4F-H} & 4F-H & 81.3 (13/16) & 16/16 & $7.51 \pm 7.77$ & $10.76 \pm 9.13$ & $2.00 \pm 1.24$ \\
        & & 4F-I & 6.3 (1/16) & 2/16 & $83.84 \pm 112.68$ & $4.74 \pm 1.78$ & $20.53 \pm 23.28$ \\
        & & 5F-I & 0.0 (0/16) & 1/16 & $178.88$ & $10.03$ & $55.32$ \\
        \midrule
        \multirow{3}{*}{Reloc3r relative-pose output} & \multirow{3}{*}{4F-H} & 4F-H & 62.5 (10/16) & 14/16 & $18.50 \pm 32.99$ & $10.39 \pm 10.61$ & $9.33 \pm 11.62$ \\
        & & 4F-I & 56.3 (9/16) & 12/16 & $12.66 \pm 20.70$ & $12.07 \pm 14.48$ & $4.22 \pm 8.61$ \\
        & & 5F-I & 75.0 (12/16) & 14/16 & $11.37 \pm 28.63$ & $14.19 \pm 18.26$ & $6.42 \pm 11.78$ \\
        \midrule

        \multicolumn{8}{c}{(d) Alignment to non-docking targets} \\
        \midrule
        GR-VIP & 4F-H & Non-docking target & N/A & 12/12 & $4.43 \pm 3.74$ & $6.69 \pm 4.69$ & $2.50 \pm 1.61$ \\
        \bottomrule
    \end{tabular*}

    \vspace{1mm}
    \parbox{\textwidth}{\scriptsize
        $n/N$: valid error samples / all attempts.
        \texttt{-}: no valid error samples.
        Standard deviation is not reported for \(n=1\).
        N/A: mechanical engagement is not applicable.
        GR-VIP results corresponding to panel (c) are reported in panels (a) and (b).
    }
\end{table*}

\subsection{Experimental Setup}
\label{subsec:experimental_setup}

\noindent\textbf{Robot Platform.}
Fig.~\ref{fig:docking_scenarios}(a) shows the mobile robot platform used in this study. The robot is equipped with two RGB cameras mounted at different heights. The upper camera provides the monocular RGB observations used as input to the policy, while the lower camera is used only for measuring terminal pose errors using a fiducial marker during evaluation and is not used for policy inference.

\noindent\textbf{Experimental Environments.}
The experiments are conducted in three indoor environments with docking stations, as shown in Fig.~\ref{fig:docking_scenarios}(d)--(f). The training environment (4F-H) is used for collecting demonstrations, while the other two environments are used to evaluate generalization. The 4F-I environment retains the same docking station as 4F-H but has a different background, whereas 5F-I differs in both the docking station and background.
All experiments are conducted in obstacle-free settings.

\noindent\textbf{Training Demonstrations.}
Training demonstrations consist of 150 successful teleoperated docking trajectories collected over approximately two hours in a single environment. 
The docking station remains visible throughout each trajectory, including those involving repositioning maneuvers to address
misalignment. The same demonstration set is used to train GR-VIP and the baseline policies evaluated under the primary and zero-shot transfer settings.

\noindent\textbf{Evaluation Setup.}
For evaluation in each environment, the robot attempts docking from the 16 different positions shown in Fig.~\ref{fig:docking_scenarios}(b), with the docking station visible at the start of each trial. Each position is evaluated with a single attempt. Among them, 14 positions are radially distributed around the target to cover different approach distances and angles. The remaining 2 positions, with a $\pm20$ cm lateral offset and a 10 cm longitudinal offset from the docking station, are configured with intentional misalignment to evaluate local realignment behavior.
During policy execution, the marker remains covered and is not used by the policy. After each execution, it is uncovered for terminal-pose measurement, as shown in Fig.~\ref{fig:docking_scenarios}(c).

\noindent\textbf{Success and Failure Criteria.}
A docking attempt is considered successful when the robot completes docking through mechanical engagement with the station. 
A repositioning maneuver is defined as an approach to the docking region followed by reversing away and initiating a new approach. A trial is considered failed if the robot performs more than five repositioning maneuvers or remains stationary for at least 40 s. Failure is also declared when an unintended collision prevents further motion or when the docking target leaves the camera field of view and cannot be recovered. Because mechanical engagement is not detected autonomously, an operator terminates each trial after confirming successful engagement or when any predefined failure criterion is met. No overall time limit is imposed.

\subsection{Baselines and Evaluation Metrics}
\label{subsec:baselines and metrics}

We compare GR-VIP with two baseline policies representing alternative visual representation strategies. As an end-to-end learned baseline, we use a goal-conditioned diffusion policy (GC-DP) adapted from NoMaD~\cite{sridhar2024nomad}, a diffusion-based policy for real-world goal-conditioned navigation. Following NoMaD, GC-DP uses an EfficientNet-B0 visual encoder trained jointly with the policy, providing a task-specific representation learned solely from the demonstration data. As a pretrained baseline, we use a DINOv3-based policy with a frozen DINOv3 encoder~\cite{simeoni2025dinov3}, which provides general-purpose visual representations learned through self-supervised pretraining. It uses the same diffusion-policy architecture as GR-VIP, differing primarily in the representation supplied to the policy.

We evaluate docking performance using two complementary metrics: docking success rate and terminal alignment accuracy. The docking success rate is computed over all 16 trials in each evaluation setting, with success defined by mechanical engagement with the docking station. Terminal alignment accuracy is quantified by the absolute lateral, longitudinal, and heading errors relative to the docking pose specified by the goal image. These errors are measured using the lower camera after each trial, and we report their mean and sample standard deviation. The statistics include successful trials and collision failures for which valid terminal-pose measurements are available. Other failure cases are excluded from the error statistics but remain included in the success-rate calculation. For non-docking targets, where mechanical engagement is not applicable, we report only terminal pose-alignment errors. Table~\ref{tab:docking_results} summarizes the quantitative results.

\begin{figure*}[!t]
    \centering
    \includegraphics[width=\textwidth]{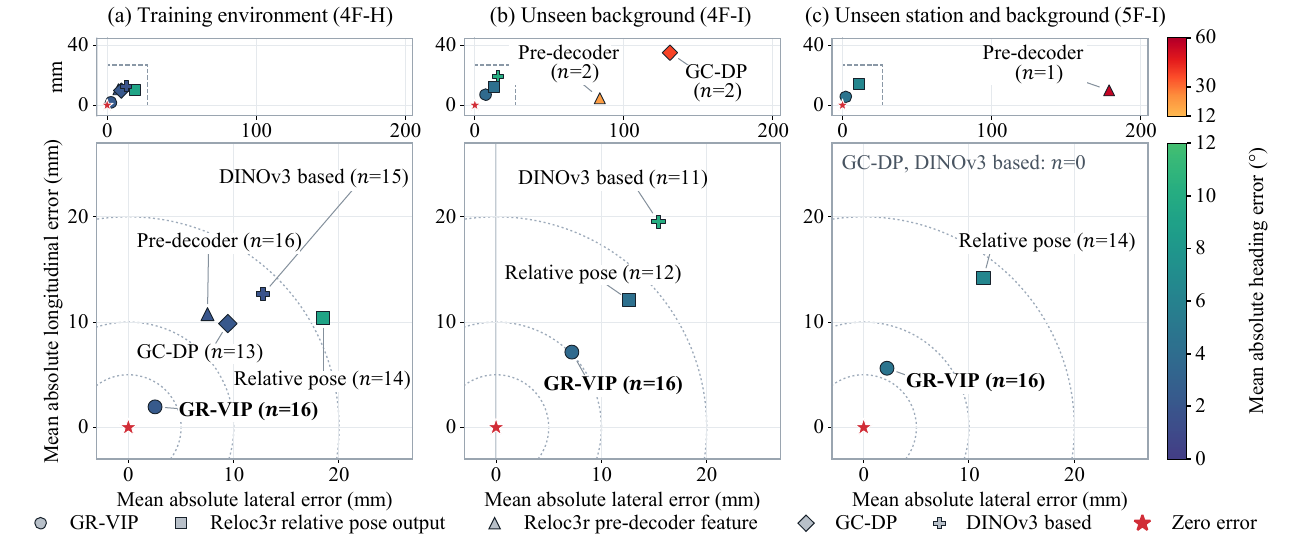}
    \caption{Overview of terminal alignment errors for policies trained on the common 4F-H demonstrations. The top row shows the full error range, while the bottom row provides a detailed view near zero error. \(n\) denotes the number of valid error samples. GR-VIP consistently exhibits low terminal alignment errors relative to the other policies across all three environments.}
    \label{fig:docking_results}
\end{figure*}

\subsection{Docking Success and Terminal Precision}
\label{subsec:docking success and terminal precision}
We first evaluate docking performance in the training environment (4F-H). As shown in Table~\ref{tab:docking_results}(a), GR-VIP succeeds in all 16 trials while maintaining millimeter-scale terminal position errors, with mean lateral and longitudinal errors of $2.53$~mm and $1.95$~mm, respectively. In contrast, GC-DP and the DINOv3-based baseline achieve success rates of 56.3\% and 50.0\%.
These results show that GR-VIP achieves more reliable precise terminal alignment than the two baseline policies in the training environment.

\subsection{Zero-Shot Generalization}
\label{subsec:zero-shot generalization}
We next evaluate zero-shot generalization by transferring GR-VIP and the two baseline policies trained on 4F-H to 4F-I and 5F-I without further training. As shown in Table~\ref{tab:docking_results}(b), GR-VIP maintains a high docking success rate in both unseen environments, achieving 100.0\% in 4F-I and 93.8\% in 5F-I, while preserving millimeter-scale terminal position accuracy. In contrast, GC-DP and the DINOv3-based baseline show substantially lower zero-shot success rates, with neither baseline succeeding in 5F-I.

To assess baseline performance when demonstrations from the evaluation environment are available, we independently train baseline models for each environment using 150 demonstrations collected in that environment. In 4F-I, the environment-specific GC-DP and DINOv3-based models achieve success rates of 31.3\% and 50.0\%, respectively, while in 5F-I they achieve 87.5\% and 81.3\%. Nevertheless, GR-VIP, trained only on 4F-H, achieves higher success rates in both unseen environments. These results demonstrate that GR-VIP transfers across changes in background and docking station without additional training.

\subsection{Representation Ablation Study}
\label{subsec:representation_ablation}
We further examine the effect of representation choice by comparing GR-VIP with variants using Reloc3r features before cross-view decoding or the final relative-pose output for policy conditioning. As shown in Table~\ref{tab:docking_results}(c), the pre-decoder feature variant performs well in the training environment, but its success rate drops sharply in the two unseen environments. The relative-pose output variant maintains moderate success across the three environments but consistently underperforms GR-VIP. Although the relative-pose output provides an explicit geometric estimate, its lower success rates suggest that the dense decoded relational tokens used by GR-VIP provide a more effective representation for policy conditioning. These results support retaining dense relational features after cross-view interaction for precise terminal alignment and transfer across environments.

For an overall comparison, Fig.~\ref{fig:docking_results} visualizes terminal alignment errors across the three evaluation environments for policies trained on the common 4F-H demonstrations. Across these environments, GR-VIP shows consistently low terminal alignment errors among the evaluated policies.

\subsection{Non-Docking Target and Low-Light Evaluation}
\label{subsec:non-docking target and low-light evaluation}

To further illustrate the goal-conditioned behavior of GR-VIP beyond physical docking, we conduct supplementary alignment experiments using four non-docking targets not included in the training demonstrations, as shown in Fig.~\ref{fig:docking_scenarios}(h)--(i). As in the docking experiments, a goal image specifies the desired relative configuration for each target, and no additional training is performed. Whereas the docking evaluation uses 16 initial poses, the non-docking setting uses three initial poses per target, each at a distance of 1.0~m and with approach angles of \(\{-30^\circ, 0^\circ, 30^\circ\}\), resulting in 12 trials in total. Table~\ref{tab:docking_results}(d) shows that GR-VIP achieves millimeter-scale terminal position errors, indicating precise goal-relative alignment on these unseen targets.

As shown in Fig.~\ref{fig:docking_scenarios}(g), we also examine GR-VIP under a low-light condition. For this experiment, we use a simplified initialization in which the robot directly faces the docking station with a heading of \(0^\circ\), at a distance of either 30~cm or 80~cm. We conduct 10 trials in total, with five trials at each distance. Under this setting, GR-VIP successfully completes all 10 docking trials, indicating that the policy is capable of docking under the low-light condition.

\section{Conclusion}
\label{sec:conclusion}
We presented GR-VIP, a goal-conditioned visuomotor policy that leverages pretrained cross-view features from relative camera pose estimation for precise terminal alignment. Real-robot experiments demonstrated high docking success rates and millimeter-scale terminal position errors in the training environment and in unseen environments without additional training. Representation ablations further highlighted the benefit of dense decoded relational tokens for policy conditioning. Supplementary experiments also illustrated the policy's ability to align with non-docking targets and complete docking under a low-light setting. Together, these results support the use of geometry-aware cross-view representations as a useful prior for precise goal-relative control, while allowing the goal image to specify the desired configuration without additional target-specific processing.

Future work will investigate precise terminal alignment under partial target occlusion and in the presence of static and dynamic obstacles. We will also explore ways to improve heading accuracy under unseen conditions and conduct broader evaluations across diverse docking stations, backgrounds, and lighting conditions.

\FloatBarrier
{
\bibliographystyle{IEEEtran}
\bibliography{reference}
}

\end{document}